\documentclass[10pt,twocolumn,letterpaper]{article}

\usepackage{cvpr}              % To produce the CAMERA-READY version

\usepackage{multirow}

\usepackage{xcolor}
\usepackage{array}

\newcolumntype{g}{>{\color{gray}}c}

\definecolor{cvprblue}{rgb}{0.21,0.49,0.74}
\usepackage[pagebackref,breaklinks,colorlinks,allcolors=cvprblue]{hyperref}
\usepackage{gensymb}
\usepackage{booktabs}

\def\paperID{} % *** Enter the Paper ID here #18197
\def\confName{CVPR}
\def\confYear{2026}

\title{KV-Tracker: Real-Time Pose Tracking with Transformers}

\author{Marwan Taher \quad Ignacio Alzugaray \quad  Kirill Mazur\quad 
Xin Kong \quad  Andrew J. Davison\\
Dyson Robotics Lab, Imperial College London\\
{\tt\small \{m.taher, i.alzugaray, k.mazur21, x.kong21, a.davison\}@imperial.ac.uk}
}

\begin{document}
\maketitle
\begin{abstract}

Multi-view 3D geometry networks offer a powerful prior but are prohibitively slow for real-time applications. We propose a novel way to adapt them for online use, enabling real-time 6-DoF pose tracking and online reconstruction of objects and scenes from monocular RGB videos. 

Our method rapidly selects and manages a set of images as keyframes to map a scene or object via $\pi^3$~\cite{wang2025pi3} with full bidirectional attention. We then cache the global self-attention block's key-value (KV) pairs and use them as the sole scene representation for online tracking. This allows for up to $15\times$ speedup during inference without the fear of drift or catastrophic forgetting. Our caching strategy is model-agnostic and can be applied to other off-the-shelf multi-view networks without retraining.

We demonstrate KV-Tracker on both scene-level tracking and the more challenging task of on-the-fly object tracking and reconstruction without depth measurements or object priors. Experiments on the TUM RGB-D, 7-Scenes, Arctic and OnePose datasets show the strong performance of our system while maintaining high frame-rates up to ${\sim}27$ FPS.

Project Page: \url{https://marwan99.github.io/kv_tracker/}

\end{abstract}
% \textbf{PITCH} Multi-view networks are powerful but prohibitvly slow for real-time application, we present a way to adapt them for online use. We further demonstrate it on the more challenging task of object tracking and reconstruction.

\section{Introduction}
\label{sec:intro}

\begin{figure}
    \centering
    \includegraphics[width=\columnwidth]{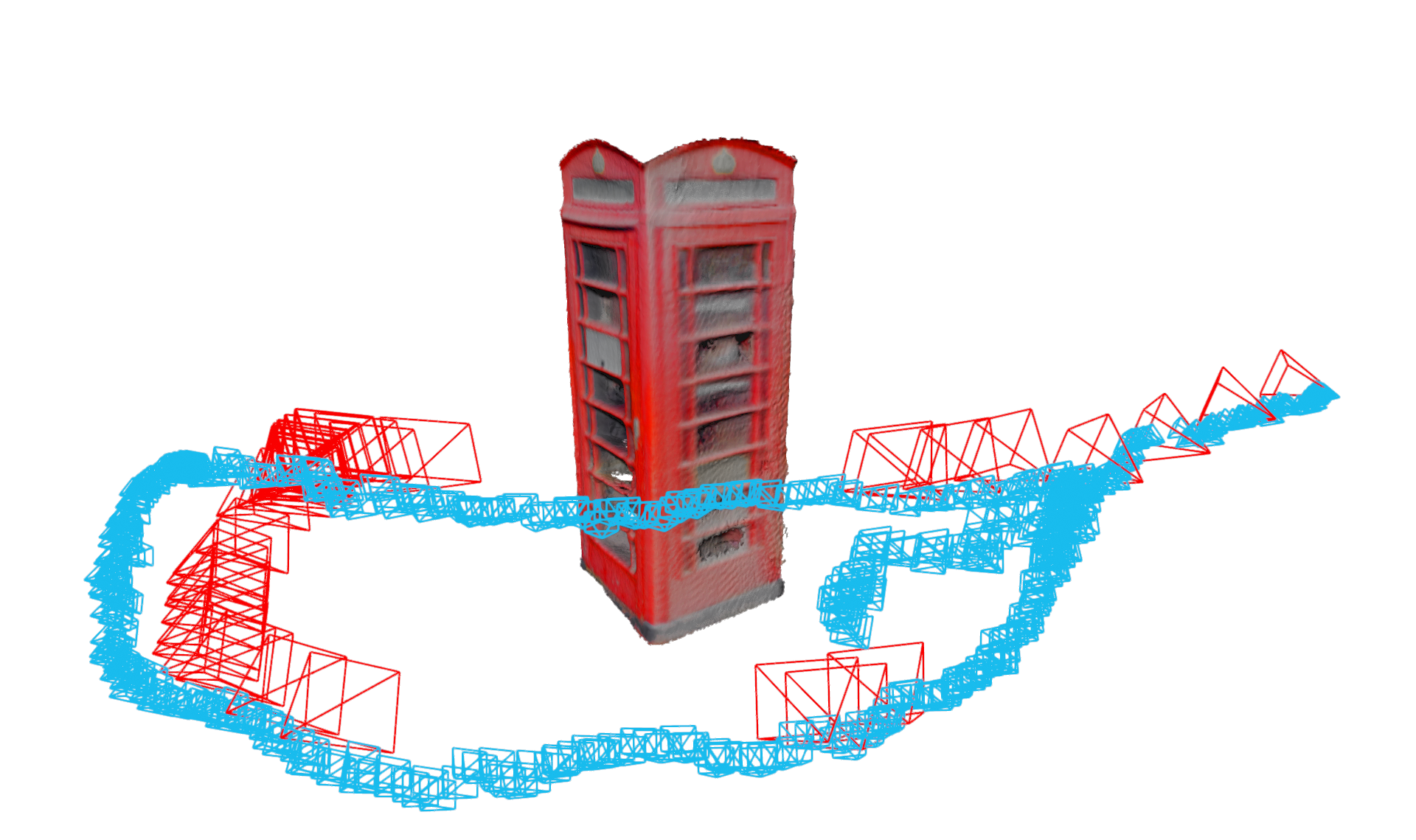}
    \caption{Real-time object tracking and online reconstruction using an RGB images from a camera scanning a telephone booth. Red frustums indicate keyframes used for mapping; blue frustums indicate tracking frames processed at ${\sim}27$ FPS. The geometry shows the global reconstruction obtained by fusing point maps from keyframes.}
    \label{fig:example}
\end{figure}

% PROBLEM
Scene reconstruction and tracking, especially at the object level, are important for applications such as AR/VR, self-driving, and robotics. While it is a well-studied topic, accurate and robust real-time 3D tracking remains challenging even if precise object shape models are available in advance. It is even more difficult when a representation of the object must also be acquired as part of the real-time process.

Multi-view 3D reconstruction models based on transformers and trained on large datasets are currently having a huge impact on 3D vision.
The rapid development of these models began with DUSt3R \cite{wang2024dust3r} which can regress point-maps from a pair of views. This work led to MASt3R \cite{leroy2024mast3r}, which provided features that can be used for explicit 2-view matching. Models with multi-view inputs emerged next, such as VGGT \cite{wang2025vggt}, $\pi^3$ \cite{wang2025pi3}, and MapAnything \cite{keetha2025mapanything}.
Since these models are attention-based, they have $O(N^2)$ computational cost in the number input frames. Pair-wise networks $(N = 2)$ have successfully been used as the front-end for real-time 3D vision systems (e.g.  MASt3R-SLAM \cite{Murai_2025_mast3r-slam}), where they provide strong calibration-free 
priors in a single pass, complementing explicit geometry-based optimisation.

However, it is more challenging to make real-time use of multi-view models ($N\geq2$), which restricts their application in real-time systems~\cite{maggio2025vggt-slam}.
These models are powerful but monolithic, making it unclear how to apply them to a continuous image stream. For instance, after running a model with 50 frames to generate a 3D reconstruction, what should one do when a new frame arrives? Run the model again from scratch with all 51 frames? This is not a scalable approach to online tracking and reconstruction.

Interesting architectures designed and trained to consume a stream of images rather than an off-line set have recently emerged. These build an implicit memory of previous observations \cite{wang20253spann3r, wang2025continuous}. While locally the output might be consistent, globally these methods suffer from drift and lack the ability to ``close the loop'' when trajectories get long and camera positions are revisited. This is particularly problematic in object tracking, where a rotating object quickly returns to previous relative camera/object poses.

% Other methods have attempted to speed up the inference of off-the-shelf multi-view models by running them in a causal manner where the estimate for each frame is only affected by the past via a causal attention mechanism. However, these methods are also doomed to drift in loopy tracking scenarios.
% \andy{any citations for causal?}

% OUR PROPOSAL
In this paper we propose \textit{KV-Tracker}, which 
uses multi-view networks to track in real-time while maintaining the rich multi-view observation from previous frames. We achieve this by extracting a set of key-value (KV) pairs from the global attention blocks and saving them as a representation of the observations during mapping inferences. During tracking, we can perform inference with a single query frame, re-localising the current observation against the saved KV representation without altering it, thus not poisoning the stored representation.

This significantly improves the runtime inference speed, enabling 27 FPS tracking, a $15\times$ speed-up over recomputing the keys and values for frames whose geometry we already know. This makes it possible to use these multi-view models in an online fashion, since the redundant, repetitive compute is significantly reduced from quadratic to linear complexity in the number of keyframes.

We showcase applications of such a method for camera tracking against a scene and zero-shot object tracking with no prior information about the object or depth, with the latter being more challenging since it has fewer visual cues than a typical full scene observation. Typically, object tracking methods have required access to object CAD models or depth measurements, which are restrictive requirements that we relax with our proposed method.

Our key contributions are the following:
\begin{itemize}
    \item A novel application of multi-view feed-forward geometry networks to real-time object and scene tracking.
    \item A new scene representation based on cached Key-Value pairs.
    \item A method for zero-shot object tracking and reconstruction on the fly with no object-specific prior information.
\end{itemize}

\section{Related Work}

\subsection{Classical Reconstruction and Representations}

% SfM \cite{schoenberger2016sfm, schoenberger2016mvs, agarwal2011building} and SLAM \cite{davison2007monoslam, klein2007ptam, mur2015orb} have traditionally relied on hand-crafted features and sparse scene reconstructions, gradually progressing towards dense reconstructions \cite{newcombe2011dtam, izadi2011kinectfusion}. More recent work has explored alternative scene representations, including code-based approaches~\cite{bloesch2018codeslam}, NeRF-based methods such as iMAP~\cite{sucar2021imap}, and Gaussian splatting~\cite{matsuki2024gaussian}. Modern 3D reconstruction models sit along this continuum, leveraging data-driven priors to handle increasingly challenging scenarios. In this work, we explore using multiview reconstruction models for tracking and adapt them for real-time operation.
% 

Traditional SLAM and reconstruction methods can be broadly categorized  by their tracking strategy. Dense methods like DTAM~\cite{newcombe2011dtam} and KinectFusion~\cite{newcombe2011kinectfusion} track against explicit  dense models of the scene, enabling robust localization. In contrast, keyframe-based methods like PTAM~\cite{klein2007ptam} and ORB-SLAM~\cite{mur2015orb} track against sparse features from selected keyframes, achieving real-time performance through reduced representation complexity.

Our approach fits within the tracking against a model paradigm. The cached key-value pairs from multi-view attention serve as a learned scene representation that encodes geometric and appearance information from keyframes.

\subsection{Multi-view Reconstruction Models}

DUSt3R~\cite{wang2024dust3r} introduced a 2-view reconstruction model that, without requiring any calibration information, predicted point maps end-to-end from pixels, with one of the frames acting as a reference frame. To enhance the estimated geometry, MASt3R~\cite{leroy2024mast3r} added a feature prediction head along with the point map prediction head for dense correspondences, which was later used by MASt3R-SfM~\cite{duisterhof2025mast3r} to solve an offline structure-from-motion problem. MASt3R-SLAM~\cite{Murai_2025_mast3r-slam} introduced a real-time SLAM system built with MASt3R as its front-end.

To overcome the pairwise limitation, 
Fast3R~\cite{wang2025faster} operated on multi-view information with global all-to-all self-attention between frames, inferring geometry from multi-view inputs in a single forward pass. VGGT~\cite{wang2025vggt} added local self-attention layers to alternate with global self-attention and trained a network with multi-task learning on other outputs, providing strong generalisation. 

$\pi^3$~\cite{wang2025pi3} adopted VGGT's network architecture but reformulated the scene geometry output to local point maps instead of global point maps, dropped the special camera register tokens, and trained a model with a permutation-invariant loss, so the model is less sensitive to the reference view choice. Following the same architecture, MapAnything~\cite{keetha2025mapanything} added optional input modalities as conditioning, such as depth and camera intrinsics, along with more decoding heads for depth maps and other modalities.

In all these multi-view networks, information sharing happens between the multi-views via global all-to-all self-attention, with the computation growing quadratically with the number of input views, causing the inference speed to drop. Addressing this limitation for streaming settings is the core motivation behind our work.

\paragraph{Streaming Models}

To address online operation on image streams, Spann3R~\cite{wang20253spann3r} processed a sliding window of images incrementally online, predicting global point maps for the input views in a shared reference frame. Unlike our approach, their memory was generated from each frame independently, rather than jointly.

CUT3R~\cite{cut3r} processed input images online via a recurrent neural network, incorporating the current view's information into the hidden state representation and updating it. Since it was trained on a maximum of 64 views, it struggled to handle long sequences, leading to drift and catastrophic forgetting. To increase robustness, TTT3R~\cite{chen2025ttt3r} added a confidence-guided state update rule via the attention map between state queries and the observation keys; this allowed for better handling of longer sequences. However, since this state is updated with every observation, TTT3R is still prone to drifting and requires state resets to operate on long sequences. Our method does not have this state corruption issue, since our KV-cache-based state representation is anchored in a set of keyframes and not updated on every new input frame.

Long3R~\cite{chen2025long3r} maintains an implicit memory that is updated via an attention-based memory gating mechanism. Kinaema~\cite{sariyildiz2025kinaema} introduced an RNN-based network that can re-localise query views against previously observed parts of the scene that are encoded into a latent memory. MUSt3R~\cite{cabon2025must3r} uses a latent memory that gets updated with sufficient viewpoint change in the online setting. This operating mode is similar to ours, though ours leverages off-the-shelf trained models and does not require extra training or fine-tuning.

StreamingVGGT~\cite{streamVGGT} trains a network with causal attention to model streaming inputs. Point3R~\cite{wu2025point3r} trains a network to run in a streaming fashion with spatial features encoded at explicit 3D locations in the scene serving as memory that gets updated and fused with new information. Similarly, our KV-cache memory is anchored in the keyframes; however, we do not require any training or fine-tuning.

% - Faster VGGT~\cite{wang2025faster} with Block-Sparse Global Attention
% FastVGGT~\cite{shen2025fastvggt} speeds up VGGT by adopting a token merging to speed up inference, 

\subsection{Object Tracking and Pose Estimation}

Object-level 6-DOF tracking is challenging due to limited visual context compared to full scenes, leading most methods to rely on priors  such as CAD models or depth sensors.

\paragraph{CAD-Based Methods} 
Many object pose estimation methods rely on having CAD models of the objects a priori. DeepIM~\cite{li2018deepim} was an RGB-based method that predicted the relative 6-DOF pose between an input image and a rendering of the object's CAD model via an iterative matching network. Having a CAD model is a strong assumption that can be limiting in different applications. Similar to ours, FoundPose~\cite{ornek2024foundpose} used off-the-shelf pretrained models for object pose estimation; however, they relied on having CAD models.
Several lines of work, such as MegaPose~\cite{labbe2022megapose} and FoundationPose~\cite{wen2024foundationpose}, followed this recipe of render-and-compare, scaling up network training by training on large synthetic datasets. These methods, however, required depth information at estimation time.

\paragraph{RGB-D Reconstruction} 
Like our method, BundleTrack~\cite{wen2021bundletrack} reconstructed objects on the fly. An object pose graph was built and solved online, recovering the poses of keyframes and using them for tracking. BundleSDF~\cite{wen2023bundlesdf} built on top of this and learned a neural signed distance field as an object representation. These methods relied on depth measurements.

\paragraph{RGB Reconstruction} 
OnePose~\cite{sun2022onepose} and OnePose++~\cite{he2022onepose++} scan objects and build offline sparse and semi-dense object reconstructions, respectively. They also trained 2D-3D matching networks through which they solve PnP and recover the object pose. These two methods are the closest to ours in the object-level setting, as they operate on RGB inputs; however, unlike our approach, their reconstruction is performed offline rather than online.

\section{Method}
\label{sec:method}

\subsection{Overview}

\begin{figure}[t]
    % \centering
    \includegraphics[width=\columnwidth]{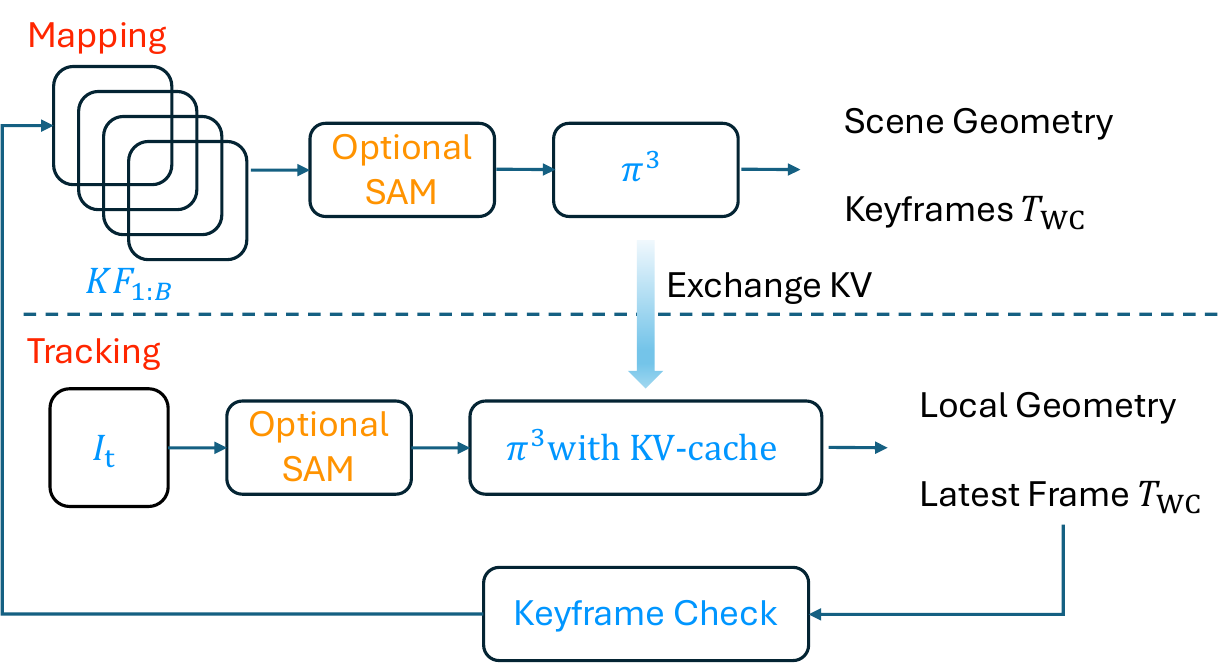}
    \caption{This is an overview of our real-time tracking and online reconstruction method. Our system is decomposed into 2 interleaved steps mapping and tracking. During the mapping stage a set of keyframes $KF_{1:B}$ are used to generate a KV-cache.  During the tracking the latest frame $I_t$ tracked using the latest available cached key-value pairs. The KV-cache is refreshed with new keyframe insertions. During object level mode, segmentation masks can be used, along with our keyframing system.}
    \label{fig:system_overview}
\end{figure}

Our objective is to use $\pi^3$ \cite{wang2025pi3} in an online fashion, building a map via a set of keyframes and tracking the latest frames in real-time as shown in Figure~\ref{fig:system_overview}. We provide an analysis of the model's architecture and present a novel way to  enable online operation on streaming images. We demonstrate this via 2 tasks: real-time camera tracking and real-time object tracking. Our method does not require any further training or fine-tuning.

Our method consists of 2 interleaved processes, that can be parallelised similar to PTAM \cite{klein2007ptam}:
\begin{itemize}
    \item \textbf{Mapping}, where a set of input images is automatically selected as keyframes, from which we obtain a KV cache as an implicit scene representation.
    \item \textbf{Tracking}, where the latest frame's $I_t$ state can be estimated in real-time against the latest KV-cache scene representation, achieving up to \textbf{27 FPS}.
\end{itemize}

\subsection{Details of $\pi^3$}
\label{sec:prelim}

$\pi^3$ is feed-forward decoder only transformer based model. Given a set of input images $I_n \in \mathbb{R}^{H \times W \times 3}, \quad n = 1,\ldots,N$, the model will predict a set camera poses in a world frame $T_n \in \mathbb{SE}\!\left(3\right)$, point maps in the local camera frames $P^c_n \in \mathbb{R}^{H \times W \times 3}$ and a confidence score per 3D point $C_n \in \mathbb{R}^{H \times W}$.

Initially the input images are patchified into $M$ patches and encoded via a ViT backbone to produce a set of tokens $\operatorname{Enc}(I_{1:N})=X_{1:N}, \quad X \in \mathbb{R}^{M \times d_k}$, where $d_k$ is the feature vector dimension.

These set of tokens are then passed through $L$ decoding layers, alternating between frame-wise self-attention layers and global self-attention layers. These feature aggregating blocks enable multi-view information exchange. In the global self-attention blocks, every input image patch can attend to all the other input patches. This is the key to producing globally consistent 3D output.

Finally, decoding heads are used to predict the final outputs, $\text{decoder\_heads}(X_{1:N})=T_{1:N}, P_{1:N}, C_{1:N}$. A key insight is that each frame's tokens $X_n$ are decoded independently from the others, which is a crucial feature needed for adapting this model for online use. 

\subsection{Processing Bottleneck}

The model uses Scaled Dot-Product Attention \cite{vaswani2017attention} for the frame-wise and global self-attention blocks. Recall that attention is computed via: 
\begin{equation} \label{eq:attn}
\text{Attention}(Q, K, V) = \text{softmax}\left(\frac{QK^T}{\sqrt{d_k}}\right)V
~,
\end{equation}
where queries $Q$, keys $K$ and values $V$ are computed via linear projection of the tokens $X$, parametrised by $\theta_l$, where $l$ is the layer index, as below:

\begin{equation} \label{eq:linear_proj}
\text{Proj}(X; \theta_l)=Q, K, V
~.
\end{equation}
The complexity of the attention computation is a function of the size of $Q$ and $K$. In the frame-wise self-attention blocks, a batch of per-frame tokens $X_{1:N}$ gets projected into $Q_{1:N} \in \mathbb{R}^{M \times d_k}$ and $V_{1:N} \in \mathbb{R}^{M \times d_k}$ which yields attention maps which are quadratic with the number of patches per frame,  giving $\mathcal{O}(M^2)$ computational complexity.

However in global self-attention block, all tokens from all frames attend to each other. More formally, all the tokens get concatenated together into $X \in \mathbb{R}^{NM \times d_k}$ and projected to $Q \in \mathbb{R}^{NM \times d_k}$ and $V \in \mathbb{R}^{NM \times d_k}$ which yields attention maps which are quadratic in the number of input images, giving $\mathcal{O}((NM)^2)$ computational complexity. As a result of this all-to-all bidirectional attention, the computation becomes prohibitively slow when the number of frames grows, which limits the application of such model in online settings.

\subsection{Tracking via Caching}

To circumvent the computational bottleneck caused by the global self-attention layer, we propose running the full model on a set of keyframes $KF_{1:B} \subseteq I_{1:N}$, where $B$ is the number of keyframes in the keyframe buffer, with full all-to-all bidirectional attention in the global self-attention blocks, while caching the computed $\tilde{K}_{1:B}^l, \tilde{V}_{1:B}^l,  l \in  \{1, \ldots, L\} $ from each global self-attention block.

\begin{figure}[t]
    % \centering
    % \includegraphics[width=\columnwidth]{figures/uni_vs_bidirectional_attn.pdf}
    \includegraphics[width=\columnwidth]{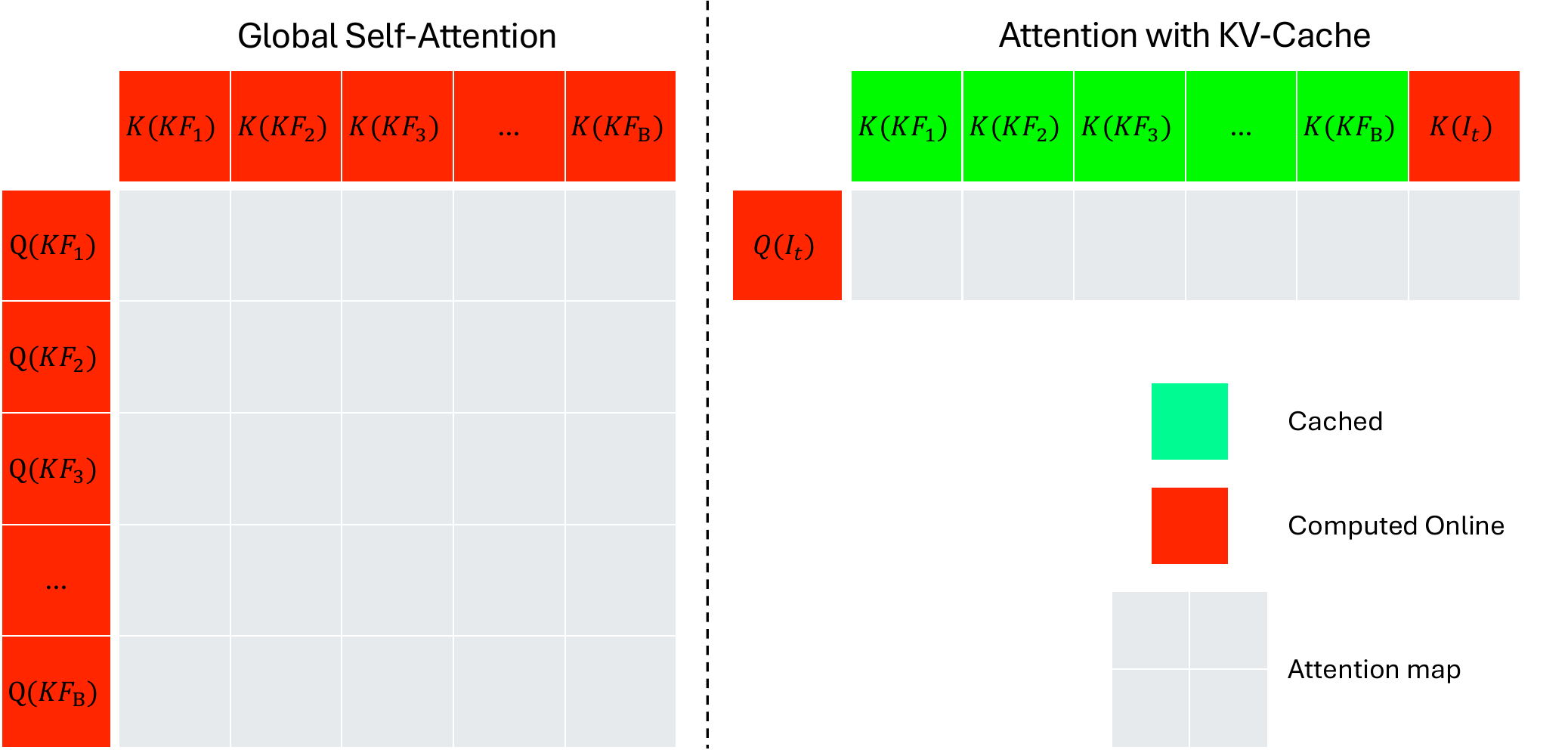}
    \caption{Visualisation of the full bidirectional global self-attention used during mapping to generate the KV-cache (left) vs. cross-attention with KV-cache + self-attention used for tracking of live frames (right).}
    \label{fig:all2all_vs_cached_attn}
\end{figure}

Then, when a new live frame $I_t$ arrives, we obtain state estimates online via the following method. We encode only $I_t$ into $X_t=\operatorname{Enc}(I_t)$, and during the feature aggregation step we run frame-wise self-attention on $X_t$ only. Then, during global self-attention we project $X_t$ obtaining $\operatorname{proj}(X_t, \theta)=Q_t, K_t, V_t$, and attend the cached keys $\tilde{K}$ and values $\tilde{K}$ via $\operatorname{Attention}(Q_t, [\tilde{K}, K_t], [\tilde{V}, V_t])$. The attention map is now obtained via:
\begin{equation} \label{eq:cached_qk}
\mathbf{Q}\mathbf{K}^T = 
\underbrace{
\begin{bmatrix}
Q_t
\end{bmatrix}
}_{M \times d_{\text{k}}}
\underbrace{
\begin{bmatrix}
\tilde{K}_{kf_1}^T & \cdots & \tilde{K}_{kf_M}^T & \tilde{K}_{t}^T
\end{bmatrix}
}_{d_{\text{k}} \times M(N+1)}
\end{equation}
This brings down the complexity from $\mathcal{O}((NM)^2)$ to $\mathcal{O}(M^2 (N+1))$ which is much cheaper as we demonstrate in our experiments. Finally, the tokens $X_t$ processed through the feature aggregator are decoded normally via $\operatorname{Decoder}(X_t)$, obtaining the camera pose and scene geometry. Figure \ref{fig:all2all_vs_cached_attn} shows the difference between the Bidirectional attention used in the mapping process and the the unidirectional attention used for tracking. 

% This unidirectional attention is similar to proposed attention structure in the perceiver architecture \cite{jaegle2021perceiver}. 

% \begin{equation} \label{eq:cached_v}
% \mathbf{V} = 
% \begin{bmatrix}
% \tilde{V}_{kf_1} & \tilde{V}_{kf_2} & \cdots & \tilde{V}_{kf_M} & V_t
% \end{bmatrix}^T
% \end{equation}

\subsection{KV-cache as a Scene Representation}

In scene reconstruction and SLAM settings, a scene gets represented by a set of primitives which can be explicit like sparse 3D points~\cite{mur2015orb} or implicit like with an MLP~\cite{sucar2021imap}. These primitives are used to encode the scene from which the scene geometry and camera poses can be recovered during mapping and tracking. 

In our formulation the KV-cache obtained via the keyframes can be interpreted as a scene representation since it was generated by exchanging multi-view information and it allows for the recover of scene geometry and camera for a given query frame.

% The cached KV tokens can be thought of as a representation of an object or scene's 3D and appearance properties, and an alternative to explicit representations such as textured meshes or point clouds. If they have been recovered from a sufficient set of keyframes around an object, this representation can be enough for robust and accurate 3D tracking over long periods of time without drift. 
% \andy{agreed that a more technical insight here would be good}

%encoding features about the object's geometry and appearance and the query tokens of the current frame contain a feature representation of the object from the current view points. The attention blocks performs the task of trying to work out where should I place these new tokens w.r.t to the keyframes tokens I already know about the scene.

\subsection{Object Tracking}

Along with using this system for scene level camera tracking, we propose applying it to the task of online object reconstruction and tracking, which is more difficult both because an object usually only fills a small fraction of the pixels in an image, and because manipulated objects often move and rotate rapidly relative to the camera. The prior information embedded our approach can be powerful in dealing with the challenges of object tracking, which also fits nicely with what tracking via caching can offer, since less than 50-60 keyframes are usually sufficient to represent a comprehensive view of an object, enabling full reconstruction of the object while maintaining high frame rate tracking.  

Real-time segmentation is now a standard computer vision component, and we make use of this in object tracking. Given segmentation masks of the object of interest, the background pixels can be masked out as black pixels. We use a simple keyframing criterion to trigger adding new keyframes. When a new keyframe is added, the KV-cache gets recomputed fully.

\paragraph{Keyframing}
We employ an angular threshold-based keyframe selection strategy to 
ensure sufficient baseline and viewpoint diversity for multi-view 
reconstruction. Specifically, a new frame $I_t$ is designated as a 
keyframe when the minimum azimuth or elevation difference between the 
current frame and all existing keyframes exceeds a threshold:
\[
\min_{kf \in KF} |\phi_t - \phi_{kf}| > \tau \quad \text{or} \quad 
\min_{kf \in KF} |\theta_t - \theta_{kf}| > \tau
~,
\]

$\phi$ and $\theta$ represent the camera's azimuth and elevation angles, while $\tau$ is the threshold angle, balancing reconstruction coverage with memory efficiency.

This view-based criterion naturally adapts to camera motion, selecting  more keyframes during viewpoint changes while avoiding redundancy during small motions or stationary periods. By comparing against all existing keyframes rather than only the most recent, our strategy prevents adding redundant views even when the camera revisits previously observed regions.

To improve robustness of the tracking and the quality of the mapping, we employ a keyframe rejection policy, where keyframes with a predicted low confidence are pruned and the system reverts back to the previous KV-cache.

\section{Experiments}

\begin{figure*}[t]
\centering
\includegraphics[width=\textwidth]{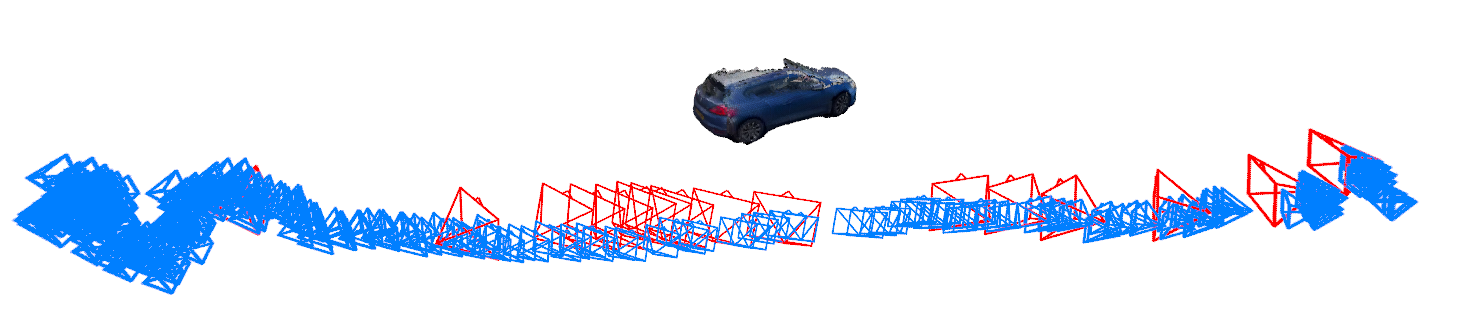}
\caption{Tracking and reconstruction demo on a moving car, captured using a phone in the wild.}
\label{fig:your-label}
\end{figure*}

In this section, we provide results on camera pose estimation~(\ref{sec:exp_cam}), object pose tracking~(\ref{sec:exp_obj}), as well as run time analysis of our system and an ablation over the effectiveness of our KV-cache self-attention scheme~(\ref{sec:exp_runtime}).

Our method adapts to any model with an architecture similar to $\pi^3$, which contains a feature aggregation blocks with global self-attention blocks~\cite{wang2025vggt,  keetha2025mapanything}. We choose $\pi^3$ since it drops the camera register token, making it less sensitive to reference views compared with VGGT. 

Since $\pi^3$ decodes the estimated camera, poses local point maps and confidence maps with 3 independent decoders, we leverage this feature and turn off the the point map and confidence decoding head, to provide speed up during tracking. Having them switched on or off only impacts the runtime performance and not the quality of tracking. To recover the scene geometry, we run with all the decoding heads enabled.

During mapping, not only do we cache the the keys and values associated with the image patch tokens, we also cache the associated per frame register tokens key value pairs. Our memory usage grows linearly with the number of keyframes used in mapping. From that it follows that our memory is also a function of the number of input images resolution, making the cache size linear with the number of patches per image.

The reported results were performed on a machine with an NVIDIA GeForce RTX 4090 graphics card, unless otherwise stated.

\subsection{Camera Tracking Evaluation}\label{sec:exp_cam}

\textbf{Baselines} We focus evaluation on other baselines that represent the scene implicitly via hidden states or memory mechanism. For that, we choose Point3R~\cite{wu2025point3r}, CUT3R~\cite{cut3r} and TTT3R~\cite{chen2025ttt3r} as our baselines, since they can run in a streaming fashion similar to ours. For a fair evaluation, we run CUT3R and TTT3R with state reset every 100 Frames, to avoid catastrophic drifting. Since LONG3R~\cite{chen2025long3r} and Kinaema~\cite{sariyildiz2025kinaema} haven't open sourced, we couldn't run their methods and compare against them.

For this task, we run our method with automatic key-framing every 50 frames. We evaluate on all the frames in the scenes. 

% All our reported metrics below are produced from the tracking branch and \textbf{not} from $\pi^3$ camera pose estimates for the keyframes.

We run our camera tracking evaluation on 7-scene and TUM RGB-D, we report standard RMSE Absolute Translation Error (ATE) metrics for the camera trajectory over the full sequences. Since our method and the baselines predict geometry up to a scale, we align all the estimated trajectories to the ground truth trajectories with a $\mathrm{Sim}\!\left(3\right)$ alignment using Umeyama Algorithm \cite{grupp2017evo}.

% \begin{table} %[h]
% \centering
% \caption{Absolute Trajectory Error (ATE) RMSE in meters on TUM-RGBD dataset.}
% \begin{tabular}{lcccc}
% \toprule
% Scene & Point3R & CUT3R & TTT3R & Ours 308 \\
% \hline
% 360   & 0.198 & 0.176 & \textbf{0.110} & 0.166 \\
% desk  & 0.337 & 0.196 & 0.104 & \textbf{0.060} \\
% desk2 & 0.355 & 0.437 & 0.147 & \textbf{0.083} \\
% plant & 0.352 & 0.383 & 0.092 & \textbf{0.048} \\
% room  & 0.947 & 0.423 & \textbf{0.253} & 0.366 \\
% rpy   & 0.056 & 0.054 & 0.054 & \textbf{0.045} \\
% teddy & 0.580 & 0.399 & 0.214 & \textbf{0.071} \\
% xyz   & 0.149 & 0.109 & 0.083 & \textbf{0.021} \\
% \hline
% Average & 0.372 & 0.272 & 0.132 & \textbf{0.108} \\
% \bottomrule
% \end{tabular}
% \label{tab:ate_tum}
% \end{table}

\begin{table} %[h]
\centering
\caption{Absolute Trajectory Error (ATE) RMSE in meters on TUM-RGBD dataset.}
\begin{tabular}{lcccgc}
\toprule
Scene & Point3R & CUT3R & TTT3R & DPVO & Ours \\
\hline
360   & 0.200 & 0.176 & \textbf{0.110} & 0.135 & 0.166 \\
desk  & 0.366 & 0.196 & 0.104 & 0.038&  \textbf{0.060} \\
desk2 & 0.321 & 0.437 & 0.147 & 0.048 & \textbf{0.083} \\
plant & 0.423 & 0.383 & 0.092 & 0.036 & \textbf{0.048} \\
room  & 0.558 & 0.423 & \textbf{0.253} & 0.394 & 0.366 \\
rpy   & 0.062 & 0.054 & 0.054 & 0.034 & \textbf{0.045} \\
teddy & 0.580 & 0.399 & 0.214 & 0.064 & \textbf{0.071} \\
xyz   & 0.134 & 0.109 & 0.083 & 0.012 & \textbf{0.021} \\
\hline
Average & 0.331 & 0.272 & 0.132 & 0.095 & \textbf{0.108} \\
\bottomrule
\end{tabular}
\label{tab:ate_tum}
\end{table}

\begin{table} %[t]
\centering
\caption{Absolute Trajectory Error (ATE) RMSE in meters on 7-Scenes dataset.}
\label{tab:ate_7scenes}
\begin{tabular}{lcccc}
\toprule
Scene & Point3R & CUT3R & TTT3R & Ours \\
\midrule
chess      & 0.427 & 0.297 & 0.154 & \textbf{0.091} \\
fire       & 0.280 & 0.218 & 0.124 & \textbf{0.042} \\
heads      & 0.389 & 0.115 & 0.097 & \textbf{0.054} \\
office     & 0.436 & 0.356 & 0.196 & \textbf{0.065} \\
pumpkin    & 0.644 & 0.249 & 0.228 & \textbf{0.142} \\
redkitchen & 0.502 & 0.118 & 0.136 & \textbf{0.038} \\
stairs     & 0.398 & 0.079 & \textbf{0.063} & 0.128 \\
\midrule
Average    & 0.439 & 0.205 & 0.143 & \textbf{0.080} \\
\bottomrule
\end{tabular}
\end{table}

We present our trajectory evaluation results for the TUM-RGBD dataset~\cite{sturm2012tum-rgbd} in Table~\ref{tab:ate_tum} and the 7-scenes dataset~\cite{shotton2013_7scene} on seq-01 in Table~\ref{tab:ate_7scenes}. Our method achieves the lowest average ATE score in both datasets, outperforming our strongest baseline TTT3R by 18\% on TUM-RGBD and by 44\% on 7-Scenes. Notably, our approach wins on 6 out of  8 scenes on TUM RGB-D and 6 out of 7 scenes on 7-Scenes, demonstrating consistent performance across diverse indoor environments.

The improvement is particularly pronounced on challenging sequences: on  TUM RGB-D's ``teddy" scene, we achieve 0.071m error compared to TTT3R's  0.214m (67\% reduction), and on 7-Scenes' ``fire'' sequence, we obtain  0.042m versus 0.124m (66\% reduction).

These results were obtained by running Point3r on 224x224, since it was running out of memory on higher resolutions. CUT3R and TTT3R were run with 512x384, to evaluate the ceiling of their performance and ours was run with 350x266. Not only can our method run at a lower resolution and outperforms all our baselines, it also does this while maintaing a higher frame rate a shown in Table~\ref{tab:fps}. In the supplementary we provide extra results running with different resolutions. we also provide more results in the supplementary with the CUT3R and TTT3R run without state resets for completeness.

\begin{table}
\centering
\caption{The estimated average FPS given the time consumed to run a single frame, for the resolutions reported.}
\begin{tabular}{lcccc}
\toprule
Method & Point3R & CUT3R & TTT3R & Ours \\
\midrule
FPS & ~5  & ~17  & ~17  & 27 \\
\bottomrule
\end{tabular}
\label{tab:fps}
\end{table}

We also included DPVO~\cite{teed2023dpvo} in our results on the TUM-RGBD dataset in Table~\ref{tab:ate_tum}. Since it is a sparse-patch odometry system we do not consider it as a baseline since it lacks dense geometry prediction. But we include it as reference since it is a strong system that can be used for tracking. Our method performs very competitively against it, with only 0.013m difference on the average ATE RMSE.

% These result show the effectiveness our approach and that anchoring our memory in the keyframes KV cache is more effective in reducing drift and improving the overall accuracy.
% To provide further granular analysis of our method, we show the distribution of our translation and rotation errors compared to TTT3R as shown in Figures~\ref{fig:ttt3r_boxplot,fig:our_boxplot}. We also provide sample trajectories from TUM-RGBD and 7-scenes in Figure~\ref{}. 

% On the "stairs" and "room" scenes where TTT3R performs comparably or 
% slightly better, we note that both methods achieve sub-0.4m accuracy. 
% Importantly, our method maintains consistent runtime performance across 
% all scenes (see Section~\ref{sec:runtime}), whereas full bidirectional 
% attention methods scale poorly with scene complexity.

% \begin{figure}
%     \centering
%     \includegraphics[width=\columnwidth]{figures/boxplot_results_ttt3r_7scenes.pdf}
%     \caption{TTT3R}
%     \label{fig:ttt3r_boxplot}
% \end{figure}

% \begin{figure}
%     \centering
%     \includegraphics[width=\columnwidth]{figures/boxplot_results_308_not_square_7scenes.pdf}
%     \caption{Ours}
%     \label{fig:our_boxplot}
% \end{figure}

\subsection{Object Tracking Evaluation}\label{sec:exp_obj}

Object tracking provides a challenging test case for evaluating how well off-the-shelf reconstruction models, trained on general-purpose datasets, generalize to object-centric scenarios. We evaluate our method in two settings: (1) comparison with other streaming reconstruction methods (CUT3R, TTT3R) on the ARCTIC dataset, and (2) comparison with methods specifically designed for object tracking (OnePose, OnePose++) on their respective benchmarks.

\begin{figure*}[t]
\centering
\includegraphics[width=\textwidth]{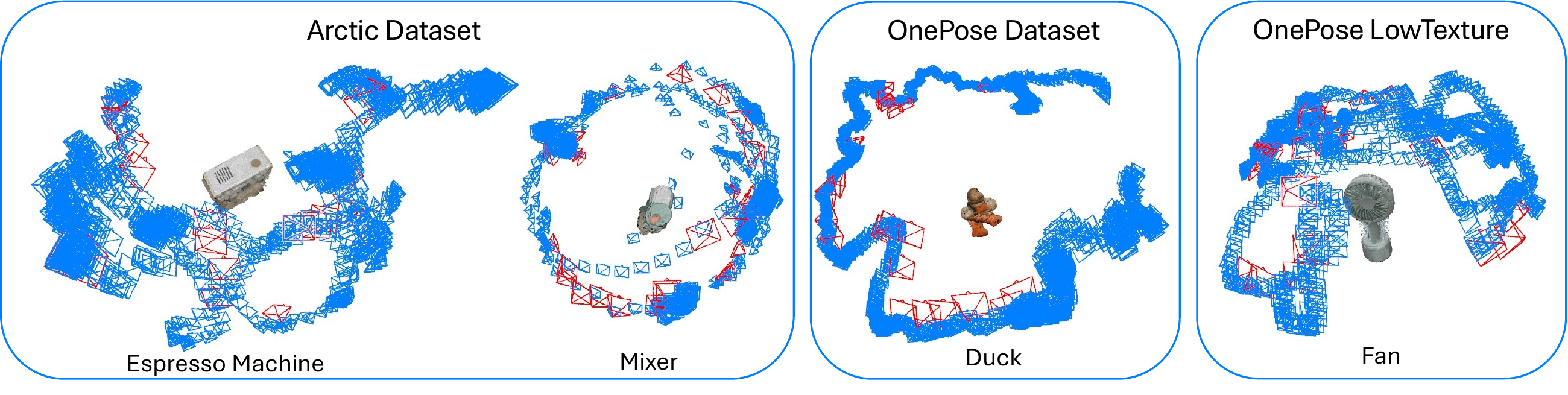}
\caption{Sample trajectory and reconstruction results from the object tracking evaluation datasets. The Arctic dataset samples are run with $308\times308$ images. While the OnePose dataset and OnePose Low Texture dataset samples are run on $518\times518$.}
\label{fig:your-label}
\end{figure*}

\textbf{Tracking protocol.} Given an initial segmentation mask of the target object, we use SAM 2~\cite{ravi2024sam} to propagate masks throughout the sequence for every input frame. Image acquisition and segmentation run in a separate process from tracking and mapping. All methods process images with backgrounds masked to black pixels. We use our object-level keyframing strategy with azimuth and elevation thresholds of $10°$ for keyframe selection.

\subsubsection{Evaluation on ARCTIC Dataset}

The ARCTIC dataset~\cite{fan2023arctic} contains sequences of various objects being manipulated by a person in tabletop settings. We use the egocentric camera view from sequences from S01 for evaluation.

We compare against CUT3R and TTT3R as baselines, running all methods on the same set of segmented frames. Table~\ref{tab:arctic_ate} shows that KV-Tracker achieves the lowest average ATE of 0.228m, outperforming both baselines (CUT3R: 0.305m, TTT3R: 0.303m). 

% Our method wins on 9 out of 10 scenes, with particularly strong improvements on challenging sequences like ``capsulemachine" (0.300m vs. 0.458m/0.552m) and ``waffleiron" (0.204m vs. 0.336m/0.342m).

Notably, none of the evaluated methods, including ours, were explicitly trained on masked images. The strong performance of our approach, which is based on $\pi^3$, suggests that the learned priors generalize well to object-centric tracking with background masking. CUT3R and TTT3R showed similar average performance to each other but exhibited greater variance across individual scenes.

\begin{table}%[t]
\centering
\small
\caption{Absolute Trajectory Error (ATE) RMSE in meters on the Arctic dataset~\cite{fan2023arctic}.}
\label{tab:arctic_ate}
\begin{tabular}{lccc}
\toprule
Scene & CUT3R & TTT3R & Ours@308  \\
\midrule
espressomachine & 0.253 & 0.175 & \textbf{0.151} \\
ketchup & 0.369 & 0.319 & \textbf{0.249} \\
microwave & 0.181 & 0.156 & \textbf{0.135} \\
box & \textbf{0.176} & 0.207 & 0.200 \\
laptop & 0.319 & 0.305 & \textbf{0.248} \\
waffleiron & 0.336 & 0.342 & \textbf{0.204} \\
scissors & 0.237 & 0.264 & \textbf{0.188} \\
capsulemachine & 0.458 & 0.552 & \textbf{0.300} \\
phone & 0.408 & \textbf{0.399} & 0.402 \\
mixer & 0.313 & 0.309 & \textbf{0.198} \\
\midrule
Average & 0.305 & 0.303 & \textbf{0.228} \\
\bottomrule
\end{tabular}
\end{table}

\subsubsection{Eval on OnePose Datasets}

OnePose~\cite{sun2022onepose} introduced a dataset of everyday objects, while its successor  OnePose++~\cite{he2022onepose++} contributed a dataset focused on low-texture objects. Ground-truth 3D bounding boxes are manually annotated, and camera poses are obtained via ARKit tracking.

Both baseline methods perform an offline reconstruction phase before tracking, providing them with complete 3D object models—information not available to our online method. Additionally, the baselines do not require object segmentation as input; instead, they leverage annotated 3D bounding boxes from the offline mapping phase, which they project into subsequent frames to estimate object locations. These bounding boxes are sometimes oversized relative to the actual objects.

We evaluate our system with segmentation masks from SAM 2~\cite{ravi2024sam}. For fair comparison, we also report results using 2D bounding boxes extracted from these masks with 50-pixel dilation to approximate the spatial extent used by the baselines.

We follow the standard evaluation protocol, reporting recall at error tolerances of 1cm–1°, 3cm–3°, and 5cm–5°. We additionally ablate our method across two input resolutions: $308 \times 308$ and $518 \times 518$ pixels.

\begin{table*}[t]
\centering
\caption{Performance comparison on OnePose and OnePose Low Texture datasets. Accuracy (\%) at different error thresholds. Baseline methods use offline 3D reconstruction.}
\label{tab:onepose_results}
\begin{tabular}{llccccccr}
\toprule
\multirow{2}{*}{Method} & \multirow{2}{*}{Input Type} & \multicolumn{3}{c}{OnePose} & \multicolumn{3}{c}{OnePose Low Texture} & \multirow{2}{*}{FPS} \\
\cmidrule(lr){3-5} \cmidrule(lr){6-8}
& & 1cm, 1° & 3cm, 3° & 5cm, 5° & 1cm, 1° & 3cm, 3° & 5cm, 5° & \\
\midrule
OnePose & 3D Bbox & 49.7 & 77.5 & 84.1 & 12.4 & 35.7 & 45.4 & 15 \\
OnePose++ & 3D Bbox & \textbf{51.1} & \textbf{80.8} & 87.7 & \textbf{16.8} & 57.7 & 72.1 & 11 \\
\midrule
Ours @ 308 & Seg. Mask & 2.25 & 47.0 & 75.7 & 1.21 & 32.2 & 62.2 & \textbf{27} \\
Ours @ 308 & 2D Bbox & 2.9 & 52.8 & 83.2 & 4.15 & 57.3 & 83.3 & \textbf{27} \\
Ours @ 518 & Seg. Mask & 10.7 & 75.5 & 92.1 & 6.85 & 62.0 & {85.7} & 16 \\
Ours @ 518 & 2D Bbox & 5.3 & 69.3 & \textbf{92.9} & 12.1 & \textbf{80.0} & \textbf{94.4} & 16 \\
\bottomrule
\end{tabular}
\end{table*}

\paragraph{OnePose Dataset} 

Despite building the object map online without any offline reconstruction phase, KV-Tracker achieves competitive performance compared to the baselines. At the most permissive 5cm, 5° threshold, our high-resolution variant (Ours@518 with 2D bounding box) achieves 92.85\% recall, outperforming both OnePose (84.1\%) and OnePose++ (87.7\%). At the intermediate 3cm, 3° threshold, our method achieves 69.29\%-75.51\% recall compared to OnePose++'s 80.8\%, a modest gap considering that the baseline methods leverage pre-built maps with known 3D structure.

The performance gap widens at the strictest 1cm, 1° threshold, where our method achieves 10.71\% recall (segmentation mask) compared to OnePose++'s 51.1\%. OnePose++ benefits from offline multi-view reconstruction that provides more complete and accurate 3D models, while our online reconstruction may have incomplete coverage during initial frames. Nevertheless, our method's ability to achieve competitive coarse-grained tracking while operating fully online at higher frame rates (16-27 FPS vs. 11-15 FPS) demonstrates a practical trade-off between accuracy and the flexibility of online operation. The reported FPS values of OnePose and OnePose++ were obtained on a V100 GPU.

% \paragraph{Onepose LowTexture Dataset}

% On the Textureless dataset our method outperfoms the baseline by a large margin on the 5cm, 5deg and 3cm, 3deg thresholds with close recall on the 1cm, 1deg of 12.06 compared to OnePose++ 12.4 with the 518 variant while keeping a higher frame rate of 16FPS vs 11 FPS. Our low resolution variant with the 2D bounding box outperforms the baselines on the 5cm, 5deg and 3cm, 3deg while running at a much higher frame rate of 29FPS. 

% The 2D bounding box variants consistently performed better than their 2D segmentation counter parts since more discriminative information is available at the object contours w.r.t to the background, since in this datasets the objects are stationary and not moving that can be advantageous in this setting especially with the low textured objects.

\paragraph{OnePose LowTexture Dataset}

Our method outperforms the baselines on the low-texture dataset at coarser thresholds. Our high-resolution variant with 2D bounding boxes achieves 94.42\% recall at 5cm, 5° (vs. OnePose++ 72.1\%) and 80.04\% at 3cm, 3° (vs. 57.7\%), while maintaining comparable performance at the strictest 1cm, 1° threshold (12.1\% vs. 16.8\%) and operating at 16 FPS compared to OnePose++'s 11 FPS. 

Notably, even our lower-resolution variant (308px with 2D bounding box) outperforms both baselines at 5cm, 5° and 3cm, 3° thresholds while running at nearly 3× the frame rate (27 FPS vs. 11 FPS). This demonstrates that our approach generalizes better to challenging low-texture scenarios than traditional feature-based methods, even without offline reconstruction.

Across both datasets, on average our variants using dilated 2D bounding boxes outperform those using segmentation masks. We attribute this to the inclusion of background context around object boundaries, which provides additional discriminative features. This effect is particularly pronounced, since the objects in these datasets are stationery while the camera is scanning them, where boundary regions contain informative visual cues.

% \begin{table*} %[t]
% \centering
% \caption{Performance comparison on OnePose and OnePose Low texture datasets. Results shown as accuracy (\%) at different distance thresholds.}
% \label{tab:performance}
% \begin{tabular}{llccccccr}
% \toprule
% \multirow{2}{*}{Method} & \multirow{2}{*}{Config} & \multicolumn{3}{c}{OnePose dataset} & \multicolumn{3}{c}{OnePose LowTexture} & \multirow{2}{*}{FPS} \\
% \cmidrule(lr){3-5} \cmidrule(lr){6-8}
% & & 1cm, 1deg & 3cm, 3deg & 5cm, 5deg & 1cm, 1deg & 3cm, 3deg & 5cm, 5deg & \\
% \midrule
% OnePose* & 3D Bounding Box  & \underline{49.7} & \underline{77.5} & 84.1 & \underline{12.4} & 35.7 & 45.4 & 15 \\
% OnePose++* & 3D Bounding Box & \textbf{51.1} & \textbf{80.8} & 87.7 & \textbf{16.8} & 57.7 & 72.1 & 11 \\
% \midrule
% % Ours@308 & default & -- & -- & -- & 0.75 & 26.6 & 54.4 & \textbf{27.8} \\
% Ours@308 & Segmentation Mask & 2.25 & 47.02 & 75.74 & 1.21 & 32.17 & 62.19 & ? \\
% Ours@308 & 2D Bounding Box & 2.87 & 52.79 & 83.22 & 4.15 & 57.33 & 83.31 & 29 \\
% Ours@518 & Segmentation Mask & 10.71 & 75.51 & \underline{92.09} & 6.85 & \underline{61.95} & \underline{85.71} & 16 \\
% Ours@518 & 2D Bounding Box & 5.31 & 69.29 & \textbf{92.85} & 12.06 & \textbf{80.04} & \textbf{94.42} & 16 \\
% \bottomrule
% \end{tabular}
% \end{table*}

% \begin{figure}
%     \centering
%     \includegraphics[width=\columnwidth]{figures/onepose.png}
%     \caption{Onepose demos place holder}
%     \label{fig:one_pose_recon_demo}
% \end{figure}

\subsection{Runtime Analysis}\label{sec:exp_runtime}

% We provide an experiment showcasing the effectiveness of our adaptation to $\pi^3$ in speeding up the inference time when using the KV-cache from the mapping instead of performing full bidirectional attention, computing new keys and values for the keyframes.

% We run a synthetic workload of image of size $308\times308$ to measure the frames per second through put when doing inference on $N$ frames, vs doing inference on $1$ frame with the KV-cache of $N$ frames. As shown in the plot in ~\ref{fig:all2all_vs_cached_attn} the computation with full bidirectional attention follows an exponential drop with the number of frames as a function of $O(N^2)$ required compute. While ours maintains 30 FPS performance for up to 50 frames, then starts to drop linearly afterwards, keeping 25 FPS for up to a cache size of 70 frames and above 20 FPS for up to 110 frames. This demonstrates the capability of our system of being able to map small workspaces or objects and being able to re-localise w.r.t them.

We conduct an experiment to demonstrate the effectiveness of our adaptation to $\pi^3$ for accelerating inference through KV-cache reuse from the mapping phase, compared to performing full bidirectional attention with fresh key-value computation for all keyframes.

\begin{figure}%[t]
    % \centering
    \includegraphics[width=\columnwidth]{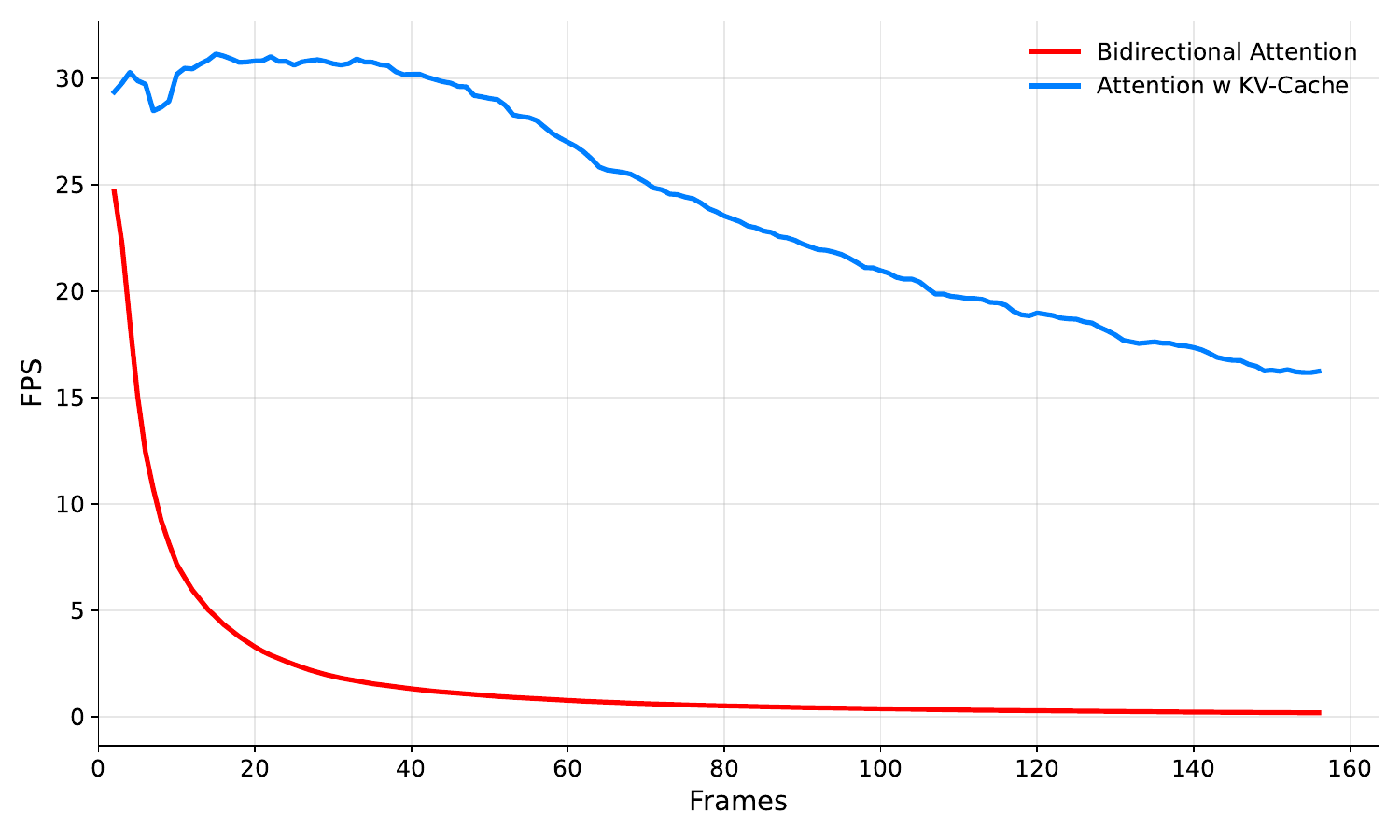}
    % \caption{Achievable frames per second (FPS) when doing inference on $N$ frames with full attention between them and inference on a single with the KV-cache of $N$ frames.}
    \caption{Frames per second (FPS) throughput comparison: processing $N$ frames with full bidirectional attention vs. processing a single query frame with KV-cache from $N$ frames.}
    \label{fig:all2all_vs_cached_attn_plot}
\end{figure}

We run a synthetic workload using images of size $308\times308$ pixels, measuring frames-per-second (FPS) throughput under two conditions: (1) full inference on $N$ frames with all-to-all attention, and (2) inference on a single query frame using cached keys and values from $N$ keyframes. As shown in Figure~\ref{fig:all2all_vs_cached_attn_plot}, the full bidirectional attention approach exhibits quadratic scaling with $O(N^2)$ complexity, resulting in rapidly degrading speed as the number of frames increases. In contrast, our KV-cache approach maintains 30 FPS for up to 50 frames, then degrades more gradually, sustaining 25 FPS with cache sizes up to 70 frames and remaining above 20 FPS for cache sizes up to 110 frames. We report results up to the point of running out of memory (24GB) during inference.

These results demonstrate that our system can efficiently map small workspaces or objects and relocalize with respect to them in real-time, maintaining practical frame rates even with substantial mapping history.

% \subsection{Ablation}

% \begin{itemize}
%     \item Run latest K frames instead of just the last one and eval accuracy \& speed
%     \item Depth degradation with rescpect to
%     \item Quantify the performance drop in full vs unidirectinoal cross attention
%     \item Highlight/plot/show how it dies very quickly after N frames for different image resolutions
% \end{itemize}

% \begin{table}[t]
% \centering
% \caption{Comparison of camera tracking performance on 7 Scenes and TUM datasets.}
% \label{tab:camera_traj_eval}
% \begin{tabular}{lcccc}
% \toprule
% \multirow{2}{*}{Method} & \multicolumn{2}{c}{7 Scenes} & \multicolumn{2}{c}{TUM} \\
% \cmidrule(lr){2-3} \cmidrule(lr){4-5}
%  & ATE & RPE & ATE & RPE \\
% \midrule
% CUT3R w reset@100 & 29.71 & \textbf{0.76} & 27.227 & 1.39 \\
% TTT3R w reset@100 & 15.36 & 0.79 & 13.23 & \textbf{1.19} \\
% Point3r src:point3r & 12.4 & 5.8 & -- & -- \\
% % Long3r & 8.72 & 5.03 & 5.4 & 2.36 \\
% Ours @ 308 & 9.14 & 1.30 & 10.60 & 3.50 \\
% Ours @ 518 & 9.23 & 1.23 & \textbf{9.88} & 2.72 \\
% Ours @ 224 & \textbf{9.07} & 1.649 & 14.409 & 4.026 \\
% \bottomrule
% \end{tabular}
% \end{table}

% \begin{tabular}{|c|g|c|}
% \hline
% Header 1 & Header 2 & Header 3 \\
% \hline
% Data 1 & Data 2 & Data 3 \\
% \hline
% \end{tabular}
\section{Conclusion}

% We presented KV-Tracker, a training free method for adapting 3D reconstruction methods for real-time operation. Our approach uses memory anchored in keyframe KV-cached pairs, we show its capabilties in camera tracking and online object-centric reconstruction and tracking. This method is suitable for small work spaces, such object, or small environments. The available memory can be a limitation to explore larger environments, investigating cache pruning and compression and efficient incremental KV-cache computation are interesting directions to explore to bring this system closer to a SLAM system.

We presented KV-Tracker, a training-free approach for adapting 3D reconstruction models to real-time camera tracking and online reconstruction. By caching key-value pairs from keyframes as memory, our method enables efficient tracking and online reconstruction of objects and small workspaces.

Our system's memory requirements currently limit its application to spatially confined environments or objects. Future work includes exploring cache pruning, compression, and efficient incremental KV computation to scale toward full SLAM systems capable of handling larger scenes. 
% \newpage

% WARNING: do not forget to delete the supplementary pages from your submission 

{
    \small
    \bibliographystyle{ieeenat_fullname}
    \bibliography{main}
}

\clearpage
\setcounter{page}{1}
\maketitlesupplementary

\section{Applying KV-Tracker on Depth Anything V3}

% To demonstrate the applicability of our proposed approach in making off-the-shelf reconstruction methods effeciect for online use. We show our tracking via caching method on Depth Anything 3~\cite{depthanything3}. Despite being a very recent model, it still use self-attention blocks for exchanging information between the input frames. We show evaluation results of running it the 7-Scenes dataset.

To demonstrate the applicability of our proposed approach in making off-the-shelf reconstruction methods efficient for online use, we show our tracking via caching method on Depth Anything 3~\cite{depthanything3}. Despite being a very recent model, it still uses self-attention blocks for exchanging information between the input frames, so our method can be applied on it. We show evaluation results of running it on the 7-Scenes and TUM RGB-D datasets.

\begin{table}[h]
\centering
\caption{Absolute Trajectory Error (ATE) RMSE in meters on 7~-Scenes dataset for Depth Anything V3, 1.15B model, running at 18 FPS.}
\begin{tabular}{lc}
\hline
 & DepthAnything \\
\hline
chess & 0.098 \\
fire & 0.044 \\
heads & 0.055 \\
office & 0.106 \\
pumpkin & 0.143 \\
redkitchen & 0.083 \\
stairs & 0.299 \\
\hline
Average & 0.118 \\
\hline
\end{tabular}
\label{tab:your_label}
\end{table}

\begin{table}[h]
\centering
\caption{Absolute Trajectory Error (ATE) RMSE in meters on TUM RGB-D dataset for Depth Anything V3, 1.15B model, running at 18 FPS.}
\begin{tabular}{lc}
\hline
 & DepthAnything \\
\hline
360 & 0.19 \\
desk & 0.173 \\
desk2 & 0.261 \\
plant & 0.096 \\
room & 0.525 \\
rpy & 0.059 \\
teddy & 0.102 \\
xyz & 0.024 \\
\hline
Average & 0.179 \\
\hline
\end{tabular}
\label{tab:your_label}
\end{table}

\section{Additional results}

We provide additional results on our baselines in the scene-level camera tracking evaluation for CUT3R~\cite{cut3r} and TTT3R~\cite{chen2025ttt3r} with different input image resolutions for completeness. At the 224x224 their frame rate increases to 22 FPS, which is still lower than our 27 FPS.

\begin{table}[h]
\caption{Absolute Trajectory Error (ATE) RMSE in meters on 7~-Scenes dataset.}
\centering
\small
\begin{tabular}{lcccccc}
\hline
 & CUT3R & TTT3R & CUT3R & TTT3R & Ours & Ours \\
Resolution & 224 & 224 & 512 & 512 & 224 & 308 \\
\hline
chess & 0.439 & 0.395 & 0.297 & 0.154 & 0.106 & \textbf{0.091} \\
fire & 0.454 & 0.256 & 0.218 & 0.124 & 0.056 & \textbf{0.042} \\
heads & 0.225 & 0.121 & 0.115 & 0.097 & 0.064 & \textbf{0.054} \\
office & 0.383 & 0.321 & 0.356 & 0.196 & 0.073 & \textbf{0.065} \\
pumpkin & 0.391 & 0.185 & 0.249 & 0.228 & 0.139 &\textbf{ 0.142} \\
redkitchen & 0.322 & 0.132 & 0.118 & 0.136 & 0.050 & \textbf{0.038} \\
stairs & 0.263 & 0.128 & 0.079 & \textbf{0.063} & 0.092 & 0.128 \\
\hline
Average & 0.354 & 0.220 & 0.205 & 0.143 & 0.083 & \textbf{0.080} \\
\hline
\end{tabular}
\label{tab:your_label}
\end{table}

\end{document}